\documentclass[letterpaper]{article} 
\usepackage{aaai24}  
\usepackage{times}  
\usepackage{helvet}  
\usepackage{courier}  
\usepackage[hyphens]{url}  
\usepackage{graphicx} 
\usepackage{natbib}  
\usepackage{caption} 
\usepackage{algorithm}
\usepackage{algorithmic}

\usepackage{amssymb}
\usepackage{autobreak}
\usepackage{multirow}
\usepackage{booktabs} 
\usepackage{adjustbox}
\newtheorem{proposition}{Proposition}
\newtheorem{thm}{Theorem}
\newcommand{\squishlist}{
 \begin{list}{$\bullet$}
  { \setlength{\itemsep}{0pt}
     \setlength{\parsep}{2pt}
     \setlength{\topsep}{2pt}
     \setlength{\partopsep}{0pt}
     \setlength{\leftmargin}{1.5em}
     \setlength{\labelwidth}{1em}
     \setlength{\labelsep}{0.5em} } }
\newcommand{\squishend}{
  \end{list}  }
\usepackage{subcaption}

\usepackage{xcolor}

\usepackage[title]{appendix}

\usepackage{newfloat}
\usepackage{listings}
\DeclareCaptionStyle{ruled}{labelfont=normalfont,labelsep=colon,strut=off} 
\floatstyle{ruled}
\newfloat{listing}{tb}{lst}{}
\floatname{listing}{Listing}
\title{Imitating Cost-Constrained Behaviors in Reinforcement Learning}
\author{
    Qian Shao\textsuperscript{\rm 1},
    Pradeep Varakantham\textsuperscript{\rm 1},
    Shih-Fen Cheng\textsuperscript{\rm 1}
}
\affiliations{
    \textsuperscript{\rm 1}School of Computing and Information Systems\\

    Singapore Management University\\
    \{qianshao.2020@phdcs.,
    pradeepv@,
    sfcheng@\}smu.edu.sg
}

\begin{document}

\maketitle

\begin{abstract}
Complex planning and scheduling problems have long been solved using various optimization or heuristic approaches. In recent years, imitation learning that aims to learn from expert demonstrations has been proposed as a viable alternative to solving these problems. Generally speaking, imitation learning is designed to learn either the reward (or preference) model or directly the behavioral policy by observing the behavior of an expert. Existing work in imitation learning and inverse reinforcement learning has focused on imitation primarily in unconstrained settings (e.g., no limit on fuel consumed by the vehicle). However, in many real-world domains, the behavior of an expert is governed not only by reward (or preference) but also by constraints. For instance, decisions on self-driving delivery vehicles are dependent not only on the route preferences/rewards (depending on past demand data) but also on the fuel in the vehicle and the time available. In such problems, imitation learning is challenging as decisions are not only dictated by the reward model but are also dependent on a cost-constrained model. In this paper, we provide multiple methods that match expert distributions in the presence of trajectory cost constraints through (a) Lagrangian-based method; (b) Meta-gradients to find a good trade-off between expected return and minimizing constraint violation; and (c) Cost-violation-based alternating gradient. We empirically show that leading imitation learning approaches imitate cost-constrained behaviors poorly and our meta-gradient-based approach achieves the best performance.
\end{abstract}

\section{Introduction}
Complex planning and scheduling problems have long been studied in the literature using a wide variety of optimization algorithms and heuristics. One generic way to think about the process of solving these complex problems is to first define a set of states, where a state represents the minimal amount of information required to describe a planning or scheduling instance; then for each state, compute the recommended action by optimizing certain well-defined performance measure. A comprehensive list of (state, action recommendation) tuples is then called the policy for the solved problem and can be used in practice easily by looking up the encountered states. 

An alternative paradigm to the above classical approach is to directly ``learn'' from an expert. More specifically, with a few expert-demonstrated traces of actions, try to generalize and derive actionable policies. Such an approach is generally called ``imitation learning'', which aims to replicate expert behaviors by directly observing human demonstrations, eliminating the need for designing explicit reward signals as in reinforcement learning (RL) \cite{abbeel2004apprenticeship}. This idea has been successfully applied in domains such as robotics \cite{fang2019survey}, autonomous vehicles \cite{kuefler2017imitating}, and game AI \cite{hussein2017imitation}.
Typically, this is achieved through techniques such as behavioral cloning \cite{bain1995framework}, inverse RL \cite{ng2000algorithms}, and generative inverse reinforcement learning (GAIL) \cite{ho2016generative}. 

The previous research in the fields of imitation learning and inverse reinforcement learning has primarily concentrated on mimicking human behaviors in unconstrained environments, such as mimicking driving a vehicle without any limitations on fuel consumption by the vehicle. However, in many practical planning and scheduling scenarios, experts consider not only rewards or preferences but also limitations or constraints. For example, the decisions made by a self-driving delivery vehicle are not only based on route preferences or rewards, which are derived from past demand data but also on the amount of fuel/power available in the vehicle. As another example, when an agent is being trained to drive a car on a race track, the expert demonstrations that the agent is mimicking involve high-speed driving and precision maneuvering, which are critical for success in a race. However, it is also essential for the agent to adhere to safety constraints, such as staying within the track's boundaries and avoiding collisions with other vehicles. These safety constraints differ from the reward function, which may focus on achieving a fast lap time or winning the race. Therefore, the agent must strike a balance between the goal of imitating the expert demonstrations and the need to adhere to the safety constraints in order to successfully complete the task. 

In scenarios where the decision-making process is influenced by both a reward model and a cost constraint model, the implementation of imitation learning becomes significantly more complex. This is because the decisions made are not solely based on the reward model, but also take into consideration the limitations imposed by the cost constraint model. To that end, we provide a new imitation learning problem in cost-constrained environments. 


Our work is closely related to \citet{malik2021inverse} and \citet{cheng2023prescribed}. In \citet{malik2021inverse}, cost constraints have to be learned from expert trajectories in scenarios where the reward function is already specified. On the other hand, in \citet{cheng2023prescribed}, both safety constraint (maximum cost limit) and cost signals are known and provided, while the reward signal remains undisclosed. 

The major differences between our proposed work and the above two past studies are that we assume cost signals (from sensors) from the environment are known and provided, but the safety constraint (maximum cost limit) is not given. In other words, we assume that we have the necessary sensors to monitor cost signals (e.g., battery level and temperature of an EV), yet we do not know the constraints on these sensor values (e.g., safety ranges on battery level and temperature).

In terms of the solution approach, our work is similar to \citet{cheng2023prescribed} in that we also utilize the combination of the Lagrangian-base method and the GAIL framework \cite{ho2016generative}. However, since we do not assume knowledge of the safety constraint, we cannot include the maximum cost limit in our objective function during training.



The major challenges we plan to address with this paper are approaches that could execute imitation learning while ensuring cost threshold constraint, which is revealed only via the observations of expert trajectories. Our key contributions are summarized as follows:
\begin{itemize}
    \item First, we formulate the cost-constrained imitation learning problem which represents the challenge of imitation learning in cost-constrained environments (where cost signals are known).
    
    \item We propose three methods to address the cost-constrained imitation learning problems. First, we design a Lagrangian-based method utilizing a three-way gradient update to solve the cost-constrained imitation learning problem. We then provide a meta-gradient approach that is able to tune the Lagrangian penalties of the first approach to significantly improve the performance. Finally, we use a cost-violation-based alternating gradient approach which has a different gradient update depending on the current solution feasibility.  
    
    \item To further validate the effectiveness of our proposed method, we conducted extensive evaluations in Safety Gym environments \cite{Ray2019} and MuJoCo environments \cite{todorov2012mujoco}. Our numerical evaluations show that the ensemble of our three proposed approaches can effectively imitate expert behavior while satisfying cost constraints.
\end{itemize}

\section{Background and Related Work}
In this section, we describe the two problem models of relevance in this paper, namely Constrained MDPs and Imitation Learning. We also briefly review related work. 

\subsection{Constrained Markov Decision Process}
Reinforcement Learning problems are characterized by an underlying Markov Decision Process (MDP), which is defined by the tuple $(\mathcal{S},\mathcal{A},\mathcal{R},\mathcal{P})$. Where $\mathcal{S}$ represents the set of states, $\mathcal{A}$ represents the set of actions. The reward function, $\mathcal{R}:\mathcal{S}\times\mathcal{A}\mapsto \mathbb{R}$, provides a quantitative measure of how well the system is performing based on the current state and action. The transition function, $\mathcal{P}:\mathcal{S} \times \mathcal{A} \times \mathcal{S} \mapsto [0,1]$, defines the probability of transitioning from one state to another, given the current state and action taken. Specifically, the probability of transitioning from state $s$ to $s'$, given that action $a$ is taken, is represented by $\mathcal{P}(s'|s, a)$. A feasible set of policies, denoted as $\Pi$, contains all possible policies that can be implemented in the system. The objective of MDP is to find an optimal policy, $\pi \in \Pi$, by maximizing the reward-based objective function, which is defined as follows:
\begin{equation}
\label{MDP obj}
\max \limits_{\pi \in \Pi} \mathbb{E}_{\pi}[\sum_{t=0}^\infty \gamma^t r(s_t,a_t)]. 
\end{equation}

In this work, we examine the scenario in which agents aim to optimize their rewards while adhering to policy-based cost constraints. This leads to an extension of the traditional MDP framework referred to as the Constrained Markov Decision Process (CMDP) \cite{altman1999constrained}. The objective in a CMDP problem is succinctly formulated as:
\begin{eqnarray}
\label{CMDP obj}
\max \limits_{\pi \in \Pi} & \mathbb{E}_{\pi}[ \sum_{t=0}^\infty \gamma^t r(s_t,a_t)] \\
s.t. & \mathbb{E}_{\pi}[ \sum_{t=0}^\infty \gamma^t d(s_t,a_t)] \leq d_0, \nonumber
\end{eqnarray}
where $d(s, a)$ is the cost associated with taking action $a$ in state $s$ and is independent of the reward function, $r(s, a)$. $d_0$ is the expected cost threshold for any selected policy. There have been numerous approaches proposed for solving Constrained MDPs~\cite{satija2020,pankayaraj2023,huy2024}, when the reward and transition models are not known {\em a priori}.

\subsection{Imitation Learning}

Methods of Reinforcement Learning require clearly defined and observable reward signals, which provide the agent with feedback on their performance. However, in many real-world scenarios, defining these rewards can be very challenging. Imitation learning, on the other hand, offers a more realistic approach by allowing agents to learn behavior in an environment through observing expert demonstrations, without the need for accessing a defined reward signal.

An effective method for addressing imitation learning challenges is Behavior Cloning (BC) \cite{bain1995framework}. This approach utilizes the states and actions of a demonstrator as training data, allowing the agent to replicate the expert's policy \cite{pomerleau1991efficient}. One of the advantages of this method is that it does not require the agent to actively interact with the environment, instead, it operates as a form of supervised learning, similar to classification or regression. Despite its simplicity, BC is known to suffer from a significant drawback: the compounding error caused by covariate shift \cite{ross2011reduction}. This occurs when minor errors accumulate over time, ultimately resulting in a significantly different state distribution.

Another approach, Inverse Reinforcement Learning (IRL) \cite{ng2000algorithms} aims to identify the underlying reward function that explains the observed behavior of an expert. Once the reward function is determined, a standard Reinforcement Learning algorithm can be used to obtain the optimal policy. The reward function is typically defined as a linear \cite{ng2000algorithms,abbeel2004apprenticeship} or convex \cite{syed2008apprenticeship} combination of the state features, and the learned policy is assumed to have the maximum entropy \cite{ziebart2008maximum} or maximum causal entropy \cite{ziebart2010modeling}. However, many IRL methods are computationally expensive and may produce multiple possible formulations for the true reward function. To address these challenges, Generative Adversarial Imitation Learning (GAIL) \cite{ho2016generative} was proposed. GAIL directly learns a policy by using a discriminator to distinguish between expert and learned actions, with the output of the discriminator serving as the reward signal. A more recent method, known as Inverse soft-Q Learning (IQ-Learn) \cite{NEURIPS2021_210f760a}, takes a different approach by learning a single Q-function that implicitly represents both the reward and policy, thereby avoiding the need for adversarial training. With their state-of-the-art performance in various applications, we designate GAIL and IQ-learn as baselines for our algorithms.

We now describe the imitation learning problem and GAIL approach here as it serves as the basis for our method. The learner's goal is to find a policy, denoted as $\pi$, that performs at least as well as an expert policy, denoted as $\pi_E$, with respect to an unknown reward function, denoted as $r(s, a)$. For a given policy $\pi \in \Pi$, we define its occupancy measure, denoted as $\rho_\pi \in \Gamma$, as~\cite{puterman2014markov} $$\rho_\pi(s, a) = \pi(a|s) \sum_{t=0}^\infty \gamma^t P(s_t = s|\pi)$$  The occupancy measure represents the distribution of state-action pairs that an agent encounters when navigating the environment with the specified policy $\pi$. It is important to note that there is a one-to-one correspondence between the set of policies, $\Pi$, and the set of occupancy measures, $\Gamma$. Therefore, an imitation learning problem can be equivalently formulated as a matching learning problem between the occupancy measure of the learner's policy, $\rho_\pi$, and the occupancy measure of the expert's policy, $\rho_{\pi_E}$. In general, the objective can be succinctly represented as the task of finding a policy that closely matches the occupancy measure of the expert's policy, which is represented as:
\begin{equation}
\label{ILobj}
\min \limits_{\pi} -H(\pi) + \psi^*(\rho_\pi - \rho_{\pi_E}), 
\end{equation}
where $H(\pi)\triangleq \mathbb{E}_\pi[-\log \pi(a|s)]$ is the causal entropy of the policy $\pi$, which is defined as the expected value of the negative logarithm of the probability of choosing an action $a$ given a state $s$, under the distribution of the policy $\pi$. Additionally, the distance measure between the state-action distribution of the policy $\pi$, represented by $\rho_\pi$, and the expert's state-action distribution, represented by $\rho_{\pi_E}$, is represented by the symbol $\psi^*$. Specifically, the distance measure (Jensen-Shannon divergence) employed by the GAIL framework is defined as follows: 
\begin{equation}
\label{GAIL distance measure}
\psi^*(\rho_\pi - \rho_{\pi_E}) = \max \limits_{D} \mathbb{E}_\pi[\log D(s,a)] + 
\mathbb{E}_{\pi_E}[\log(1-D(s,a))]
\end{equation}

The GAIL method utilizes a combination of imitation learning and generative adversarial networks, where $D \in (0,1)^{\mathcal{S} \times \mathcal{A}}$ acts as the discriminator. Through this formalism, the method trains a generator, represented by $\pi_\theta$, to generate state-action pairs that the discriminator attempts to distinguish from expert demonstrations. The optimal policy is achieved when the discriminator is unable to distinguish between the data generated by the generator and the expert data.

In our problem, as we aim to address the imitation learning problem within the constraints of an MDP, we have employed a unique distance measure that diverges from the traditional GAIL framework. This approach allows us to more effectively navigate the complexities of the constrained MDP setting and achieve our desired outcome.

\section{Lagrangian Method}

In this section, we first describe the problem of cost-constrained imitation learning and outline our approach to compute the policy that mimics expert behaviors while satisfying the cost constraints. 

We work in the $\gamma-$discounted infinite horizon setting, and we denote the expected reward and cost in association with policy $\pi \in \Pi$ as: $\mathbb{E}_{\pi}[r(s,a)]  \triangleq \mathbb{E}_{\pi} [\sum_{t=0}^\infty \gamma^t r(s_t,a_t)]$ and $\mathbb{E}_{\pi}[d(s,a)]  \triangleq \mathbb{E}_{\pi} [\sum_{t=0}^\infty \gamma^t d(s_t,a_t)]$, where $s_0 \sim p_0, a_t \sim \pi(\cdot|s_t)$, and $s_{t+1} \sim \mathcal{P}(\cdot|s_t, a_t)$ for $t \geq 0$. Formally, the Cost-Constrained Imitation Learning problem is a combination of the CMDP problem~\eqref{CMDP obj} and the Imitation Learning problem~\eqref{ILobj}, and can be characterized as:
\begin{eqnarray}
\min_{\pi} & -H(\pi) + \psi^*(\rho_{\pi} - \rho_{\pi_E}) \nonumber \\
\quad  s.t. \quad  & \mathbb{E}_{\pi}[d(s,a)] \leq \mathbb{E}_{\pi_E}[d(s,a)]
\end{eqnarray}
The imitation learning objective includes two terms: one maximizes the policy's entropy, ensuring all feasible alternatives are considered; the other minimizes the disparity between occupation measures of the policy and expert trajectories. The constraint ensures that the expected cost using the policy must be less than that of expert trajectories. We assume that $d(s,a)$ is known, but the reward is not.

Our generative approach to computing the policy that mimics the behavior of the expert within cost constraints relies on computing a solution to the unconstrained objective function of Equation~\eqref{obj} below. The theoretical justification for choosing this objective function is provided in the Appendix, and our analysis is based on the GAIL framework. Equation~\eqref{obj} is composed of three optimizations: 
\squishlist
  \item Minimize the distance between state, action distributions of the new policy, $\pi_\theta$, and expert policy, $\pi_E$. This is transformed into the loss associated with a discriminator, $D_{\omega}$, which discriminates between (state, action) pairs from experts, and (state, action) pairs generated by the new policy $\pi_\theta$.

  \item Maximize the entropy of the new policy, $\pi_\theta$, to ensure none of the correct policies are ignored. 

  \item Minimize the cost constraint violations corresponding to the new policy, $\pi_\theta$.
\squishend
\begin{eqnarray}
\label{obj}
 & L(\omega,\lambda,\theta) \triangleq \min \limits_{\theta} \max \limits_{\omega,\lambda}  \mathbb{E}_{\pi_\theta}[\log D_\omega(s,a)] + \nonumber \\ 
 & \mathbb{E}_{\pi_E}\left[\log(1-D_\omega(s,a))\right] + \nonumber \\
 & \lambda \left( \mathbb{E}_{\pi_\theta}[d(s,a)] - \mathbb{E}_{\pi_E}[d(s,a)] \right) - \beta H(\pi_\theta) ,
\end{eqnarray}
where $\theta$ represents the parameters of the policy, $\beta$ is the parameter corresponding to the causal entropy (since we maximize entropy similar to imitation learning) and finally, $\lambda$ is the Lagrangian multiplier corresponding to the cost constraint. 
$H(\pi_\theta)\triangleq \mathbb{E}_{\pi_\theta}[-\log \pi_\theta(a|s)]$ is the casual entropy of policy $\pi_\theta$.The given expert policy is represented by $\pi_E$, and a known cost function, represented by $d$, is also incorporated into the objective function. 

Given the three optimization components, we do not choose one of the three but instead compute a saddle point $(\theta, \omega, \lambda)$ for \eqref{obj}. To accomplish this, we employ a parameterized policy, represented by $\pi_\theta$, with adjustable weights $\theta$, as well as a discriminator network, represented by $D_\omega$, which maps states and actions to a value between 0 and 1, and has its own set of adjustable weights $\omega$. The Lagrangian multiplier, denoted by $\lambda$, is for penalizing the number of cost constraint violations.

To obtain the saddle point, we update the parameters of policy, discriminator, and Lagrangian multiplier sequentially: \\
\noindent \textbf{Updating $\omega$:}
The gradient of \eqref{obj} with respect to $\omega$ is calculated as:
\begin{equation}
\label{gradient of discriminator}
\begin{aligned}
 & \bigtriangledown_\omega L(\omega,\lambda,\theta) = \mathbb{E}_{\pi_\theta}[\bigtriangledown_{\omega}\log D_\omega(s,a)] + \\
& \hspace{0.05in} \mathbb{E}_{\pi_E}[\bigtriangledown_{\omega} \log(1-D_\omega(s,a))]
 \end{aligned}
\end{equation}
We utilize the Adam gradient step method \cite{kingma2014adam}  on the variable $\omega$, targeting the maximization of \eqref{obj} in relation to $D$.

\noindent \textbf{Updating $\theta$:}
To update policy parameters, we adopt the Trust Region Policy Optimization (TRPO) method \cite{schulman2015trust}. The theoretical foundation of the TRPO update process involves utilizing a specific algorithm to improve the overall performance of the policy by optimizing the parameters within a defined trust region:
\begin{equation}
\label{TRPO}
\begin{aligned}
& \theta_{k+1} = \arg \max \limits_{\theta} \mathcal{L}(\theta_k, \theta), \;\;  s.t. \quad \bar{D}_{KL}(\theta || \theta_k) \leq \delta.
\end{aligned}
\end{equation}

The key challenge in applying the TRPO update process is the computation of the surrogate advantage, denoted by $\mathcal{L}(\theta_k, \theta)$. It is a metric that quantifies the relative performance of a new policy $\pi_{\theta}$ in comparison to an existing policy $\pi_{\theta_k}$, based on data collected from the previous policy:
\begin{equation}
\label{surrogate advantage}
\begin{aligned}
    & \mathcal{L}(\theta_k, \theta) = 
    & \mathop{\mathbb{E}} \limits_{s,a \sim \pi_{\theta_k}} \left[{
    \frac{\pi_{\theta}(a|s)}{\pi_{\theta_k}(a|s)} (A_r^{\pi_{\theta_k}}(s,a) - \lambda A_d^{\pi_{\theta_k}}(s,a))
    }\right]
\end{aligned}
\end{equation}
We do not have a reward function to compute the advantage and hence we utilize the output of the discriminator, represented by $-\log D_\omega(s,a)$, as the reward signal. Subsequently, we employ the Generalized Advantage Estimation (GAE) method outlined in \citet{schulman2015high} to calculate the advantage of the reward, $A_r^{\pi_{\theta_k}}(s, a)$. Additionally, we also calculate the advantage pertaining to cost, denoted as $A_d^{\pi_{\theta_k}}(s, a)$, by utilizing the GAE method, as we know the cost function.

The average KL-divergence, represented by $\bar{D}_{KL}(\theta || \theta_k)$, between policies across states visited by the previous policy, can be computed as: 
\begin{equation}
\label{kl divergence}
\begin{aligned}
\bar{D}_{KL}(\theta || \theta_k) = \mathop{\mathbb{E}} \limits_{s \sim \pi_{\theta_k}}{
    D_{KL}\left(\pi_{\theta}(\cdot|s) || \pi_{\theta_k} (\cdot|s) \right)
}
\end{aligned}
\end{equation}

\noindent \textbf{Updating $\phi_r,\phi_d$:}
We do an Adam gradient step on $\phi_r$(reward value network parameters), $\phi_d$ (cost value network parameters)to minimize the mean-squared error of reward value and cost value, as we minimize these two loss functions:
\begin{equation}
\label{gradient of reward, cost value }
\begin{aligned}
&  \min \limits_{\phi_r} \mathop{\mathbb{E}} \limits_{s \sim \pi_{\theta_k}} (V^r_{\phi_r}(s_t)-\hat{R}^r_t)^2 \\
& \min \limits_{\phi_d} \mathop{\mathbb{E}} \limits_{s \sim \pi_{\theta_k}} (V^d_{\phi_d}(s_t)-\hat{R}^d_t)^2\\
\end{aligned}
\end{equation}

Here $\hat{R}^r_t, \hat{R}^d_t $ are the reward to go and cost to go, which are calculated by the GAE method.

\noindent \textbf{Updating $\lambda$:}
We do an Adam gradient step on $\lambda$ to increase \eqref{obj}, the gradient of \eqref{obj} with respect to $\lambda$ is calculated as:
\begin{equation}
\label{gradient of lambda}
\begin{aligned}
& \bigtriangledown_\lambda L(\omega,\lambda,\theta)  = (\mathbb{E}_{\pi_\theta}[d(s,a)] -\mathbb{E}_{\pi_E}[d(s,a)])\\
\end{aligned}
\end{equation}

Algorithm~\ref{Algo: CCIL} shows the pseudocode for our approach, Cost-ConstraIned Lagrangian (CCIL). 


\captionsetup[algorithm]{labelfont=bf}
\begin{algorithm}[!tb]
\caption{Cost-ConstraIned Lagrangian (CCIL)}
\label{Algo: CCIL}
\textbf{Input}: initial parameters of policy $\theta$, reward value network $\phi_r$, cost value network $\phi_d$, discriminator network $\omega$, batch size $K$, a set of expert trajectories $\Phi_E = \{\tau_E \sim \pi_E\}$, initial Lagrangian multipliers $\lambda$, entropy parameter $\beta$, learning rates $\alpha_r,\alpha_d,\alpha_\lambda,\alpha_\omega$. \\
\textbf{Output}: Optimal policy $\pi_\theta$ 
\begin{algorithmic}[1] 
\STATE Compute the average cost of expert trajectories: $J_E= \frac{1}{|\Phi_E|}\sum_{\tau \in \Phi_E} \sum_{t=1}^T d_t$
\FOR {$k = 1,2,...$}
\STATE Collect set of learner's trajectories $\Phi_k = \{\tau_i \}$ by running policy $\pi_{\theta_k}$ for $K$ time steps.
\STATE Collect the reward $r_t$ of $K$ time steps by using the discriminator output:$r_t = -\log (D_\omega(s_t,a_t))$ 
\STATE Compute $V^r_{\phi_r}(s_t)$ and $V^d_{\phi_d}(s_t)$ of $K$ time steps.
\STATE Compute the reward and cost advantage $A^r(s_t,a_t)$ and $A^d(s_t,a_t)$, reward to go $\hat{R}^r_t$ and cost to go $\hat{R}^d_t$ of $K$ time steps by using GAE.
\STATE Compute the average episode cost of learner's trajectories: $J_k= \frac{1}{|\Phi_k|} \sum_{\tau \in \Phi_k} \sum_{t=1}^T d_t$
\STATE Update policy by using TRPO rule: \\
\resizebox{0.9\columnwidth}{!}{ $\theta' = \arg \max \limits_{\theta} \sum_{t=1}^K \frac{\pi_{\theta}(a_t|s_t)}{\pi_{\theta_k}(a_t|s_t)}(A^r(s_t,a_t)-\lambda A^d(s_t,a_t)) + \beta H(\pi_{\theta_k})$}
\STATE Update reward value network:\\
       $\phi_r' \gets \phi_r - \frac{1}{K}  \sum_{t=1}^K \alpha_r \bigtriangledown_{\phi_r} (V^r_{\phi_r}(s_t)-\hat{R}^r_t)^2 $
\STATE Update cost value network:\\
       $\phi_d' \gets \phi_d -  \frac{1}{K}  \sum_{t=1}^K \alpha_d \bigtriangledown_{\phi_d} (V^d_{\phi_d}(s_t)-\hat{R}^d_t)^2 $
\STATE Update discriminator network: \\
       $\omega' \gets \omega +  \frac{1}{K} \sum_{t=1}^K \alpha_\omega (\bigtriangledown_{\omega} [\log(D_{\omega}(s_t,a_t))] + $\\
       $\bigtriangledown_{\omega} [\log(1- D_{\omega}(s_t,a_t))] )  $
\STATE Update Lagrangian multipliers: \\
       $\lambda' \gets \lambda + \alpha_\lambda (J_k - J_E)$
\STATE $ \theta \gets \theta'$, $\phi_r \gets \phi_r'$, $\phi_d\gets \phi_d'$, $\omega \gets \omega'$, $ \lambda \gets \lambda'$.
\ENDFOR 
\end{algorithmic}
\end{algorithm}

\section{Meta-Gradient for Lagrangian Approach}
In this section, we introduce a meta-gradient approach to improve the Lagrangian method introduced in the previous section by applying cross-validation to optimize the Lagrangian multipliers. We call this approach MALM.

Meta-gradient is a strategy designed for the optimization of hyperparameters, such as the discount factor and learning rates in Reinforcement Learning problems. This approach involves the simultaneous execution of online cross-validation while pursuing the optimization objective of reinforcement learning, such as the maximization of expected return \cite{xu2018meta,calian2020balancing}. The goal is to optimize both inner and outer losses. The update of parameters on the inner loss is to update the parameters of the policy. The outer loss can be based on measures such as policy gradient loss and temporal difference loss, and is optimized by updating hyperparameters \cite{calian2020balancing} in constrained RL problems, and using Distributed Distributional Deterministic Policy Gradients (D4PG) \cite{barth2018distributed} algorithm framework. The critic loss is used in \citet{barth2018distributed} as the outer loss function to optimize the hyperparameters.

Instead of optimizing the hyperparameters, the key idea of MALM is to update the Lagrangian multiplier such that there is a better balance between reward maximization and cost constraint enforcement. Specifically, we use the outer loss that is defined as follows:
\begin{equation}
\label{outer loss}
\begin{aligned}
L_{outer}(\lambda)= \mathbb{E}_{\pi_\theta} (A^r(s,a) - \lambda d(s,a))^2
\end{aligned}
\end{equation}

Every batch is divided into training and validation data sets. The parameter update equations for the training data set are the same as described in the Lagrangian-based method (Equations \eqref{TRPO} -- \eqref{gradient of lambda}), and for the validation data set, we update the Lagrangian multiplier by minimizing the above outer loss function. MALM is similar to Algorithm \ref{Algo: CCIL} except that from lines 7 to 12, we update policy parameters, discriminator parameters, and Lagrangian multipliers based on the training data set. We also have an additional procedure to update the Lagrangian multipliers by minimizing the outer loss function based on the validation data set (see the Appendix for detailed pseudocode).


\section{Cost-Violation-based Alternating Gradient}
We now describe our third method, Cost-Violation-based Alternating Gradient (CVAG), which does not rely on Lagrangian multipliers. Like previous methods, this method also maintains a policy network, $\theta$, reward value network, $\phi_r$, and cost value network, $\phi_d$. The key novelty of this approach is in doing a feasibility check-based gradient update that is fairly intuitive.  If the average episode cost of the learner does not exceed the average episode cost of experts (cost constraint), then we update the policy parameters towards the direction of maximizing the return, which would be the following equation in TRPO:
\begin{equation}
\label{surrogate advantage for maximizing the return}
\begin{aligned}
    & \mathcal{L}(\theta_k, \theta) = 
    & \max \limits_\theta \mathop{\mathbb{E}} \limits_{s,a \sim \pi_{\theta_k}} \left[{
    \frac{\pi_{\theta}(a|s)}{\pi_{\theta_k}(a|s)} A_r^{\pi_{\theta_k}}(s,a)
    }\right]
\end{aligned}
\end{equation}

Otherwise, we update the policy towards the direction of minimizing the costs. 
\begin{equation}
\label{surrogate advantage for minimizing the cost}
\begin{aligned}
    & \mathcal{L}(\theta_k, \theta) = 
    & \min \limits_\theta \mathop{\mathbb{E}} \limits_{s,a \sim \pi_{\theta_k}} \left[{
    \frac{\pi_{\theta}(a|s)}{\pi_{\theta_k}(a|s)} A_d^{\pi_{\theta_k}}(s,a)
    }\right]
\end{aligned}
\end{equation}

The detailed pseudocode is provided in the Appendix.

\section{Experiments}
\begin{table*}[!t]
    \centering
    \fontsize{9}{10}\selectfont
    \begin{tabular}{lrrrrrr}
       \toprule
       \multirow{2}{*}{Task}  & \small{Observation} & \small{Action}  & \small{Dataset} & \multirow{2}{*}{Reward} & \multirow{2}{*}{Cost} & \small{Safety} \\
              & \small{Space}   & \small{Space} & \small{Size} &         &   &  \small{Coefficient} \\
       \midrule
       PointGoal1 & 60 & 2 &10 & 18.77 $\pm$ 4.64 & 51.1$\pm$ 3.36 & NA\\
       PointButton1 & 76 & 2 &10 & 18.56$\pm$ 3.83 & 93.5 $\pm$ 7.8 & NA \\
       CarGoal1 & 72 & 2 &10 & 25.73 $\pm$2.44 & 45.2 $\pm$ 6.35 & NA\\
       CarButton1 & 88 & 2 &10 & 11.84$\pm$ 1.36 & 196.6 $\pm$ 25.44 & NA\\
       DoggoGoal1 & 104 & 12 &10 & 4.62$\pm$ 1.55 & 57.9 $\pm$ 9.46 & NA\\
       DoggoButton1 & 120 & 12 &10 & 3.32$\pm$ 1.2 & 181.7 $\pm$ 15.48 & NA\\
       HalfCheetah-v3 & 17 & 6 &10 & 4132.72$\pm$132.82 & 547.5 $\pm$ 8.1 & 0.4\\
       Hopper-v3 & 11&  3 &10 & 3594.04$\pm$1.78 & 433.6 $\pm$ 2.24 & 0.001\\
       Ant-v3 & 27 & 8 &10 & 1263.43$\pm$142.14 & 653.2 $\pm$ 78.82 & 2\\
       Swimmer-v3 & 8 & 2 &10 & 110.47$\pm$1.01 & 63.8 $\pm$ 1.08 & 1\\
       Walker2d-v3 & 17 & 6 &10 & 1781.42$\pm$ 19.06 & 130.1 $\pm$ 21.1 & 1\\
       Humanoid-v3 & 376 & 17 &10 & 1744.53$\pm$257.22 & 206.6 $\pm$ 29 & 0.2\\
       \bottomrule
    \end{tabular}
    \caption{Environments and expert trajectories.}
   \label{expert trajectories statitics}
\end{table*}

\begin{table*}[!t]
    \centering
    \fontsize{9}{10}\selectfont
    \begin{tabular}{llrrrrrrr}
        \toprule
        Task& &BC & GAIL &IQ-learn & LGAIL& CCIL & MALM & CVAG \\
        \midrule
        \multirow{3}*{CarGoal1}  &$R_{pen}$   & 0.54$\pm$0.03 & -0.86$\pm$ 0.34  &  -0.57$\pm$ 0.5  & -0.28$\pm$0.62& 0.17$\pm$0.6 &  \textbf{0.67$\pm$ 0.08}  & 0.61$\pm$0.06 \\
        ~ & $R_{rec}$ & 53.91   &  73.57  & 0   &72.06 & \textbf{76.02} & 67.39  & 61.13\\
        ~ &Cost-Vio & \textbf{0}&  20.55  & \textbf{0}  & 7.28 & 1.54 &   \textbf{0}  & \textbf{0}  \\ 
        \midrule

        \multirow{3}*{CarButton1}  &$R_{pen}$   & 0.35$\pm$0.45 & -0.59$\pm$ 0.07  & 0.05$\pm$ 0.02 & -0.36$\pm$0.47 & 0.13$\pm$0.53 &  0.01$\pm$0.52 & \textbf{0.42$\pm$0.08} \\
        ~ & $R_{rec}$ & 56.84   &  \textbf{60.73}   &5.49   &59.29  & 58.19& 51.94 & 42.48  \\
        ~ &Cost-Vio & 0.07&  18.01    & \textbf{0}  & 13.28 & 1.32 & 13.29 &  \textbf{0} \\
        \midrule

        \multirow{3}*{PointGoal1}  &$R_{pen}$   &0.81$\pm$0.03 &0.35$\pm$ 0.78  & -1.07$\pm$ 0.91 & 0.1$\pm$0.72 &0.07$\pm$ 0.73& \textbf{0.95$\pm$0.09} &  0.87$\pm$0.11 \\
        ~ & $R_{rec}$ &80.87  &  95.05   & 2.82  & \textbf{95.15} &93.55& 94.78 & 87.32   \\
        ~ &Cost-Vio &  \textbf{0} & 7.32    & 20.3 & 8.99 &9.58 & \textbf{0} & \textbf{0}  \\ 
        \midrule
        \multirow{3}*{PointButton1}  &$R_{pen}$   &  -0.19$\pm$0.53 &-0.55$\pm$ 0.06  & -1.26$\pm$ 0.21 & 0.37$\pm$0.54 & -0.19$\pm$0.54 &  \textbf{0.44$\pm$0.53} &0.27$\pm$0.48 \\
        ~ & $R_{rec}$ & 69.5   &  \textbf{73.49}  & 7.44  &68.53  & 69.88 & 72.95&55.98  \\
        ~ &Cost-Vio & 5.19 &  15.41    & 19.71  & 3.46& 5.15& 0.87 & 1.64 \\ 
        \midrule
        
        \multirow{3}*{DoggoGoal1}  &$R_{pen}$   & -0.3 $\pm$0.57 &-0.94$\pm$0.03  & 0$\pm$ 0.01 & 0.25$\pm$0.04 & \textbf{0.36$\pm$0.07} & 0.29 $\pm$ 0.09 & -0.25$\pm$ 0.6   \\
        ~ & $R_{rec}$ & 27.92   &   \textbf{44.81}  & 0   & 25.32 &35.5 & 28.79 &  33.33   \\
        ~ &Cost-Vio & 1.69 &  14.97    & \textbf{0}  & \textbf{0} & \textbf{0} & \textbf{0}  & 1.95 \\ 
        \midrule
        
        \multirow{3}*{DoggoButton1}  &$R_{pen}$   & 0.35$\pm$0.05 & -0.38$\pm$ 0.55  & -0.01$\pm$0.04 & 0.14$\pm$0.11 & \textbf{0.35$\pm$0.06} &  0.3$\pm$ 0.05  & 0.06$\pm$ 0.1  \\
        ~ & $R_{rec}$ & 34.94   &   \textbf{41.27}    & 0   & 14.16 & 34.64 & 30.42 & 5.72  \\
        ~ &Cost-Vio & \textbf{0} &  11.73  & \textbf{0}  & \textbf{0} & \textbf{0} & \textbf{0} &\textbf{0} \\  
        \midrule
        \multirow{3}*{HalfCheetah-v3}  & $R_{pen}$   & -6.16$\pm$0.08   &-5.91$\pm$0.05 &-7.08$\pm$0.1 &-0.29 $\pm$ 1.63&0.44$\pm$0.37 &\textbf{ 0.81$\pm$0.07} &0.53 $\pm$ 0.26 \\
        ~ &$R_{rec}$ & 70.61 &  \textbf{95.19}    & 0   & 51.91 & 44.47  & 81.49  &  52.91 \\
        ~ &Cost-Vio & 392.94&  392.12   & 417.49  & 1.08  & \textbf{0} & \textbf{0} & \textbf{0}  \\
        \midrule
        
        \multirow{3}*{Hopper-v3}  &$R_{pen}$  & 0.16$\pm$0.02   & 0.48$\pm$0.61  & 0.07$\pm$ 0.03 & 0.02 $\pm$0.49&\textbf{ 1$\pm$0.01} & 0.75$\pm$0.48 & 0.5$\pm$0.59  \\
        ~ &$R_{rec}$ & 16.11 &   98.78  & 6.54&  \textbf{99.92}  & 99.71 &99.62 & 98.58 \\
        ~ &Cost-Vio & \textbf{0} &  8.59  & \textbf{0} & 7.89  & \textbf{0} & 0.62 & 0.8  \\
        \midrule

         \multirow{3}*{Ant-v3}  &$R_{pen}$   &-0.73$\pm$0.18   & -0.69$\pm$0.02  &0.15$\pm$ 0.07 & 0.77$\pm$ 0.1  & 0.06$\pm$0.61 &\textbf{0.86$\pm$0.03 }&0.53$\pm$ 0.47 \\
        ~ &$R_{rec}$ & 86.64  &   \textbf{95.59}  & 14.8  & 76.82  &79.86 & 85.76 &78.1 \\
        ~ &Cost-Vio & 213.94 &  242.25    & \textbf{0}  &  \textbf{0}  & 10.64 & \textbf{0} & 4.71  \\
        \midrule

        \multirow{3}*{Swimmer-v3}  &$R_{pen}$   & -2.37$\pm$2.02  &-0.17 $\pm$0.56  & 0.36 $\pm$ 0.05 & -0.42$\pm$ 0.12  & \textbf{ 0.7 $\pm$ 0.49} & 0.46 $\pm$ 0.6 & 0.63$\pm$ 0.49 \\
        ~ &$R_{rec}$ & 58.44   &  \textbf{94.87}    & 35.71  &   93.35 &94.76  & 94.29  &  87.41  \\
        ~ &Cost-Vio & 93.13 &  8.63   &\textbf{0} & 7.96 &0.51 &0.37 & 0.05  \\
        \midrule
        
        \multirow{3}*{Walker2d-v3}  &$R_{pen}$   & -0.34$\pm$0.06 & -0.06$\pm$0.53  &  0.19$\pm$0.06 & 0.2$\pm$0.63 & 0.73$\pm$0.5 & 0.72$\pm$0.51 &  \textbf{0.97$\pm$0.01} \\
        ~ & $R_{rec}$ & 99.2   &   \textbf{99.35}   & 19.04  &98.57  &98.53 & 98.58 & 97.25   \\
        ~ &Cost-Vio & 14.02 &  10.22   & \textbf{0} & 7.01 & 1.33 & 2.33 & \textbf{0}  \\ 
        \midrule
        \multirow{3}*{Humanoid-v3}  &$R_{pen}$   & 0.32$\pm$0.01   & -0.26$\pm$0.02  & 0.21$\pm$0.02 & 0.86 $\pm$ 0.02 & \textbf{0.95 $\pm$ 0.02} &0.39$\pm$0.52 & 0.82$\pm$ 0.03 \\
        ~ &$R_{rec}$ & 32.21   &   \textbf{122.94}   &  20.6  & 86.01 & 94.83  & 95.77  & 82.4   \\
        ~ &Cost-Vio & \textbf{0} &  73.54    & \textbf{0}  &\textbf{0} & \textbf{0} & 2.87& \textbf{0}  \\
        \bottomrule
    \end{tabular}
    \caption{Overall performance of different environments. Normalized penalized return $R_{pen}$ captures the trade-off between achieving higher rewards and making the episode cost go below the expert's episode cost, higher is better. Recovered return $R_{rec}$ evaluates how closely the agent imitates the expert's behavior, higher $R_{rec}$ means the agent imitated the expert better. Cost-Vio captures the extent the agent's episode cost goes beyond the expert's episode cost, lower is better.}
    \label{result statistics}
\end{table*}

\begin{table*}[!tb]
    \centering
    \fontsize{9}{10}\selectfont
    \begin{tabular}{p{2.5cm}p{6.5cm}p{5cm}p{2cm}}
       \toprule
       Cost type & Description & Tasks & Best methods \\
       \midrule
       Hazards & Penalizing for entering dangerous areas. & PointGoal1, CarGoal1, DoggoGoal1 & MALM \\
       Hazards + Buttons & Penalizing for hazards and pressing wrong buttons & PointButton1,CarButton1,DoggoButton1 & MALM \\
       Control Cost & Penalizing for taking excessively large actions & HalfCheetah-v3, Hopper-v3 & MALM, CCIL \\
       Control + Contact Costs &  Penalizing for having large actions and external contact force & Ant-v3, Humanoid-v3 & MALM, CCIL \\
       Speed Limit &  Penalizing for moving at a much higher speed. & Swimmer-v3, Walker2d-v3 & CVAG \\
       \bottomrule
    \end{tabular}
    \caption{Summarizing performances based on the type of cost constraints.}
   \label{tbl:perf_summary_cost_type}
\end{table*}




\begin{figure*}[!tb]
  \centering
  \includegraphics[width=7in]{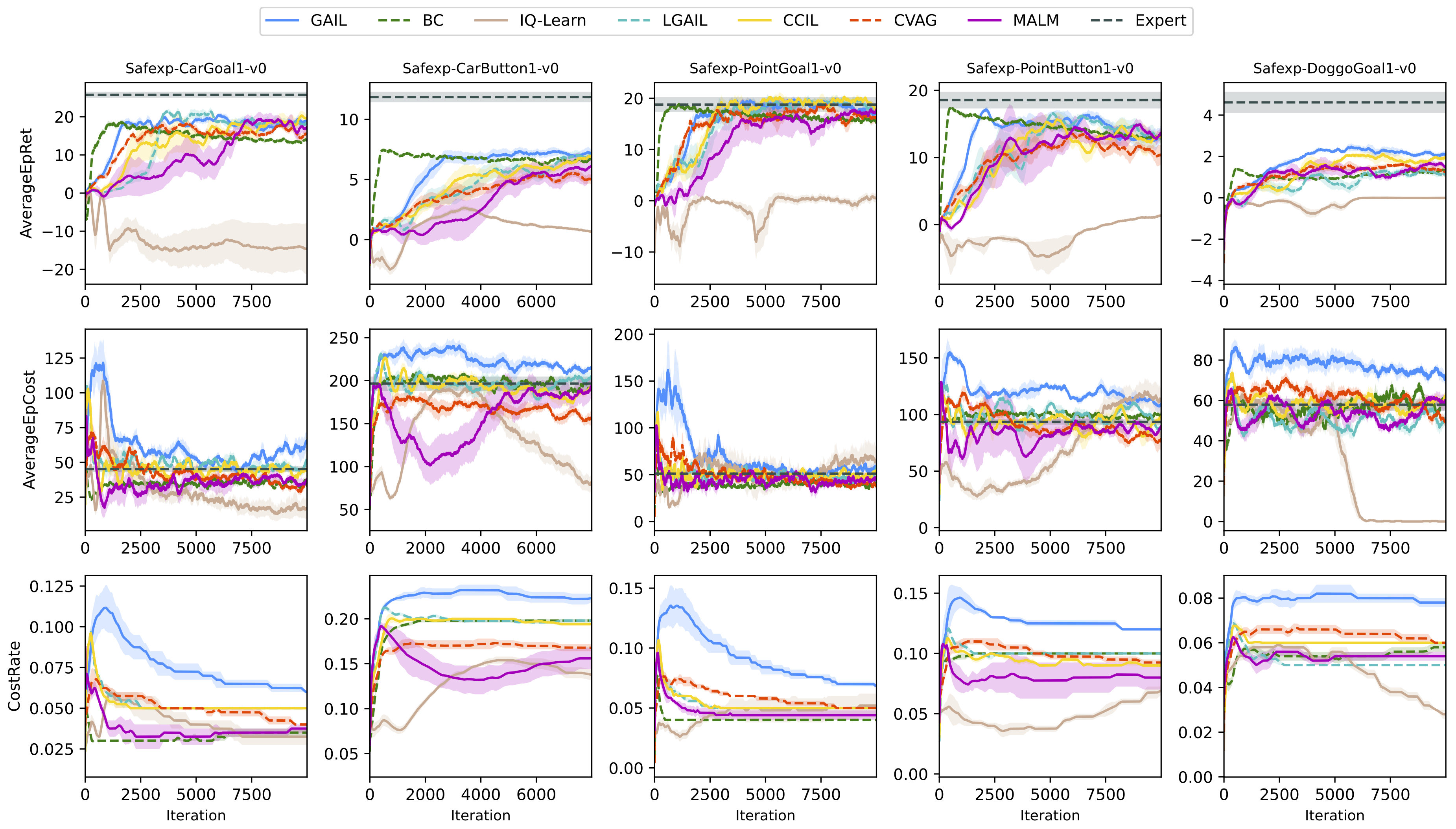}
  \caption{Performance of Safety Gym environments. The x-axes indicate the number of iterations, and the y-axes indicate the performance of the agent, including average rewards/costs/cost rates with standard deviations.}
  \label{Result Safety}
\end{figure*}


In this section, we compare our approaches against leading approaches for imitation learning (IL), including GAIL \cite{ho2016generative}, IQ-Learn \cite{NEURIPS2021_210f760a}, Behavioral Cloning (BC) \cite{bain1995framework}, and LGAIL \cite{cheng2023prescribed} in cost-constrained environments. This is to illustrate that a new approach is needed to mimic expert behaviors when there are unknown cost constraints. As we will soon demonstrate, all baselines suffer from either extensive cost constraint violations or low reward.

\subsection{Setup}

\noindent \textbf{Environments and Cost Definition}. We selected Safety Gym \cite{Ray2019} and MuJoCo \cite{todorov2012mujoco}, two well-known environments from the literature for our evaluation purpose.

The Safety Gym environment is a standard platform for evaluating constrained RL algorithms. Each instance of this environment features a robot tasked with navigating a cluttered space to achieve a goal, all while adhering to specific constraints governing its interactions with objects and surrounding areas. In our experiments, there are three robotic agents (Point, Car, and Doggo) and two task types (Goal and Button), resulting in a total of six unique scenarios. The difficulty level for these environments is standardized to 1. Throughout each timestep, the environment provides distinct cost signals for various unsafe elements, each linked to an associated constraint. Additionally, an aggregate cost signal is provided, encapsulating the overall impact of the agent's interactions with unsafe elements. The cost functions are straightforward indicators, evaluating whether an unsafe interaction has occurred ($c_t = 1$ if the agent engages in an unsafe action, otherwise $c_t = 0$).

In order to illustrate the robustness of our methods, we also adopt MuJoCo for its wide array of continuous control tasks, such as Walker2d and Swimmer. These tasks are frequently employed to assess RL and IL algorithms. For the reward, we utilized the default MuJoCo environment settings, however, as the MuJoCo environment does not have built-in cost constraints like in Safety Gym, we introduce constraints on features of the state space and action space. Table \ref{expert trajectories statitics} shows the descriptions of cost and corresponding tasks in both MuJoCo and Safety Gym environments.



In Table \ref{expert trajectories statitics}, the \textit{safety coefficient} is the cost threshold for the cost indicator. At each time step, if the cost indicator is larger than the \textit{safety coefficient}, the agent will get a cost of 1, otherwise, the cost is 0. The expert trajectories are generated by solving a forward-constrained RL problem and we summarize the statistics in the table.

\noindent \textbf{Baselines and Code}. In order to evaluate the performance of our algorithms, we compare them against three popular IL methods without explicit cost constraints consideration (GAIL, IQ-Learn, and BC). For the constraint-aware IL method, we use LGAIL as the baseline. As noted earlier, LGAIL assumes the knowledge of cost limit $d'$, which is used as a parameter in their objective function:
\begin{eqnarray}
\label{lgail obj}
 & \min \limits_{\theta} \max \limits_{\omega,\lambda}  \mathbb{E}_{\pi_\theta}[\log D_\omega(s,a)] + 
 \mathbb{E}_{\pi_E}\left[\log(1-D_\omega(s,a))\right] \nonumber \\
 & + \lambda \left( \mathbb{E}_{\pi_\theta}[d(s,a)] - d' \right) - \beta H(\pi_\theta).
\end{eqnarray}
As they defined $d'$ to be less than the minimal episode cost of expert trajectories, we use the $90\%$ of minimal episode cost of expert trajectories as $d'$. In the case of BC, the expert trajectories dataset, which consists of state-action pairs, was divided into a 70\% training data set and a 30\% validation data set. The policy was then trained using supervised learning techniques. Then in the GAIL method, the policy network, reward value network, cost value network, and discriminator network all employ the same architectures, comprising two hidden layers of 100 units each, with $\tanh$ nonlinearities being utilized in the layers. Lastly, for the IQ-Learn method, we use the same setting as illustrated in \citet{NEURIPS2021_210f760a}, that we use critic and actor networks with an MLP architecture with 2 hidden layers and 256 hidden units.

\noindent \textbf{Implementation}. We employ a neural network architecture consistent with the one utilized in the GAIL method. In addition, our approaches add a cost value network and a Lagrangian penalty term $\lambda$ (in CCIL and MALM), which distinguishes our method from GAIL. The policy, value, and cost value network are optimized through gradient descent with the Adam optimizer \cite{kingma2014adam}. The initial value of $\lambda$ is set to 0.01 and optimized using the Adam optimizer. In MALM, the state-action pairs of each batch size are divided into a 70\% training data set and a 30\% validation data set. We run each algorithm for 5 different random seeds. The algorithms ran for 2000 time steps (batch size) during each iteration, and the episode's total true reward and cost were recorded. The details of the hyper-parameters used in the experiments can be found in the Appendix. The implementation of all code is based on the OpenAI Baselines library \cite{baselines}, and can be found at \url{https://github.com/SHAOQIAN12/CCIL}.

\noindent \textbf{Performance Metrics}. We use different performance metrics to compare overall performance. Firstly, followed by \cite{Ray2019}, we record the average episode's true reward, the average episode's cost, and the cost rate over the entirety of training (the sum of all costs divided by the total number of environment interaction steps) throughout the training. We also incorporate the normalized \textit{penalized return} introduced by \cite{calian2020balancing}. This metric effectively captures the delicate balance between maximizing rewards and ensuring that the episode cost remains below the expert's episode cost. The formulation is represented as $R_{pen} = R/R_E - \mathcal{K} \max(J^\pi_d / J^{\pi_\mathbb{E}}_d - 1, 0)$, where $R$ and $J^\pi_d$ denote the average episode true reward/cost for the algorithm upon convergence (computed as an average over the last 100 iterations), and $R_E$, $J^{\pi_\mathbb{E}}_d$ represent the average episode reward/cost of the expert. The second term in the equation introduces a penalty if the episode cost exceeds the expert cost; otherwise, this term remains zero. The constant $\mathcal{K}$ serves as a fixed parameter determining the weight assigned to the constraint violation penalty. To effectively penalize algorithms consistently breaching cost constraints during evaluation, we set $\mathcal{K} = 1.2$. The \textit{recovered return} metric captures the degree to which the agent replicates the expert's behavior, denoted as $R_{rec} = R/R_E \times 100$. A value of $R_{rec} \geq 100$ signifies proficient imitation of the expert, while an approach to 0 indicates the agent's inability to replicate the expert's behavior (with $R_{rec}$ being 0 when returns are negative).Finally, \textit{cost violation} is defined as $\phi_d = \max(0, J^\pi_d - J^{\pi_\mathbb{E}}_d)$. In cases where the agent's episode cost is less than the expert's episode cost, denoting no cost violation, $\phi_d$ is set to 0.

\subsection{Results}
Table \ref{result statistics} compares performance across methods using \textit{penalized return}, \textit{recovered return}, and \textit{cost violation}. Figure \ref{Result Safety} plots training dynamics\footnote{Including average episode true reward ($R$), average episode cost ($J^\pi_d$), and cost rate ($\sum_{t=1}^N d_t / N$, where $N$ is the total number of environment interaction steps).} for most Safety Gym environments, with further details on MuJoCo environments and DoggoButton in the Appendix.
We summarize our findings in Table \ref{tbl:perf_summary_cost_type}, and elaborate our key observations below:
\squishlist
\item MALM performs best on average in Safety Gym tasks. Although there is no clear winner for tasks with hazards plus button constraints, MALM demonstrates better training performances in reward and cost overall on average according to Figure \ref{Result Safety}.

\item In MuJoCo environments focusing on control cost and contact cost, MALM leads for HalfCheetah and Ant, while CCIL excels in Hopper and Humanoid tasks. Especially in HalfCheetah, MALM closely approximates expert behavior, outperforming all competitors.

\item CVAG performs well in constraints related to speed, ranking best for Walker2d and second for Swimmer.

\item Among methods that neglect cost constraints, BC and IQ-learn have difficulty emulating expert behaviors in terms of both costs and rewards. Meanwhile, GAIL can achieve rewards close to expert levels but falls short in adhering to cost constraints, frequently incurring the highest cost rates for most scenarios.

\item LGAIL comes close to expert rewards in some tasks, performance-wise it is similar to CCIL, but often incurs higher costs throughout the training process.
\squishend

In summary, MALM shows the best balance between reward and cost adherence, with CVAG and CCIL next in line, outperforming baselines on reward and cost metrics.

\section{Conclusion}
In this study, we tackle the challenge of cost-constrained imitation learning with three scalable and effective methods. Our first approach is based on the Lagrangian relaxation. We then propose a meta-gradient technique that tunes Lagrangian penalties to enhance the performance. Finally, we propose an alternating gradient approach that adjusts gradients based on solution feasibility. Our experiments show these methods effectively imitate expert behaviors while meeting cost constraints, outperforming methods that ignore costs. The meta-gradient method strikes the best balance between high rewards and cost satisfaction.


\section{Acknowledgments}
This research is supported by the Ministry of Education, Singapore, under its Social Science Research Thematic Grant (MOE Reference Number: MOE2020-SSRTG-018).
Any opinions, findings and conclusions, or recommendations expressed in this material are those of the authors and do not reflect the views of the Ministry of Education, Singapore.

\newpage

{
\fontsize{9.5pt}{10.5pt} \selectfont

}

\clearpage


\begin{appendices}

\renewcommand\thefigure{A.\arabic{figure}}
\renewcommand\thetable{A.\arabic{table}}
\renewcommand\thealgorithm{A.\arabic{algorithm}}
\renewcommand\theequation{A.\arabic{equation}}

\setcounter{figure}{0} 
\setcounter{table}{0} 
\setcounter{algorithm}{0} 
\setcounter{equation}{0} 

\section{A. Theoretical Analysis}

The objective function of the imitation learning problem can be represented using \eqref{ILobj2}, 
\begin{equation}
\label{ILobj2}
\min \limits_{\pi} -H(\pi) + \psi^*(\rho_\pi - \rho_{\pi_E}), 
\end{equation}

And the distance measure in the GAIL framework is defined as \eqref{GAIL distance measure2}. 
\begin{equation}
\label{GAIL distance measure2}
\psi^*(\rho_\pi - \rho_{\pi_E}) = \max \limits_{D} \mathbb{E}_\pi[\log D(s,a)] + \mathbb{E}_{\pi_E}[\log(1-D(s,a))]
\end{equation}

Our proof is based on the GAIL framework, and the objective function of the cost-constrained imitation learning problem is formulated in \eqref{obj2}.
\begin{eqnarray}
\label{obj2}
 & L(\omega,\lambda,\theta) \triangleq \min \limits_{\theta} \max \limits_{\omega,\lambda}  \mathbb{E}_{\pi_\theta}[\log D_\omega(s,a)] + \nonumber \\ 
 & \mathbb{E}_{\pi_E}\left[\log(1-D_\omega(s,a))\right] + \nonumber \\
 & \lambda \left( \mathbb{E}_{\pi_\theta}[d(s,a)] - \mathbb{E}_{\pi_E}[d(s,a)] \right) - \beta H(\pi_\theta) ,
\end{eqnarray}

However, it is important to note that the form of the distance measure will differ from that of \eqref{GAIL distance measure2}, as will be explained in the following theory.

\begin{thm}
\label{Theo 1}
The objective function of the cost-constrained imitation learning problem is:
\begin{equation}
\label{CMDP gail obj}
\begin{aligned}
&\min \limits_{\pi \in \Pi} -H(\pi)+ \psi^*(\rho_\pi - \rho_{\pi_E}),
\end{aligned}
\end{equation}
where $\psi^*(\rho_\pi - \rho_{\pi_E}) = \max \limits_{D,\lambda} \mathbb{E}_\pi[\log(D(s,a))]+ \mathbb{E}_{\pi_E}[\log(1-D(s,a))] + \lambda (\mathbb{E}_\pi[d(s,a)] -\mathbb{E}_{\pi_E}[d(s,a)]) $
\end{thm}

\noindent There are broadly two steps to the proof :
\squishlist
 \item[Step 1:] Typically, optimal policy in an imitation learning setting is obtained by first solving the Inverse Reinforcement Learning (IRL) problem to get the optimal reward function $r^*$ and then running an RL algorithm on the obtained reward function. In GAIL, these two steps were compressed into optimizing a $\psi-$regularized objective. \textbf{Our first step is to show this can be also done for Cost Constrained Imitation Learning problems, albeit with an altered $\psi-$ regularized objective. }
 \item[Step 2:] Our second step is to derive the specific form of $\psi^*$ for CCIL problems. 
\squishend

\subsection{Step 1}
Constrained Markov Decision Process (CMDP) is commonly solved by utilizing the Lagrangian relaxation technique \cite{tessler2018reward}. Then CMDP is transformed into an equivalent unconstrained problem by incorporating the cost constraint into the objective function:
\begin{equation}
\label{CMDP lagrangian}
\max \limits_{\lambda \geq 0} \min \limits_{\pi \in \Pi} \mathbb{E}_{\pi}[-r(s,a)] + \lambda(\mathbb{E}_{\pi}[d(s,a)] - d_0)
\end{equation}

In the aforementioned equation, our objective is to find the saddle point of the minimax problem. Since the reward function $r(s, a)$ is not provided, our goal is to determine the optimal policy by utilizing the expert policy $\pi_E$ and the given cost functions $d(s, a)$. To accomplish this, we utilize the maximum casual entropy Inverse Reinforcement Learning (IRL) method \cite{ziebart2008maximum,ziebart2010modeling} to solve the following optimization problem:
\begin{equation}
\label{IRL}
\begin{aligned}
\max \limits_{r \in \mathcal{R} \atop \lambda \geq 0} \left( \min \limits_{\pi \in \Pi} -H(\pi) + \mathbb{E}_{\pi}[-r(s,a)] + \lambda(\mathbb{E}_{\pi}[d(s,a)] - d_0) \right) \\
\hspace{0.1in} - \left(\mathbb{E}_{\pi_E}[-r(s,a)] + \lambda(\mathbb{E}_{\pi_E}[d(s,a)] - d_0) \right) \\
\end{aligned}
\end{equation}

Where $\mathcal{R}$ is a set of reward functions. Maximum casual entropy IRL aims to find a reward function $r\in \mathcal{R}$ that gives low rewards to the learner's policy while giving high rewards to the expert policy. The optimal policy can be found via a reinforcement learning procedure:

\begin{equation}
\label{RL}
RL(r,\lambda) = \mathop{\arg \min} \limits_{\pi \in \Pi} -H(\pi)+\mathbb{E}_{\pi}[\lambda d(s,a)-r(s,a)] -\lambda d_0
\end{equation}

We study policies generated through reinforcement learning, utilizing rewards learned through IRL on the most extensive set of reward functions, denoted as $\mathcal{R}$ in Eq.\eqref{IRL}, which encompasses all functions mapping from $\mathbb{R}^{\mathcal{S} \times \mathcal{A}}$ to $\mathbb{R}$. However, as the use of a large $\mathcal{R}$ can lead to overfitting in the IRL process, we employ a concave reward function regularizer \cite{finn2016guided}, denoted as $\psi$, to define the IRL procedure:
\begin{equation}
\label{IRL regu}
\begin{aligned}
& IRL_\psi(\pi_E,d) = \\
&\hspace{0.1in}\mathop{\arg\max}\limits_{r \in \mathbb{R}^{\mathcal{S}\times \mathcal{A}} \atop \lambda \geq 0} 
\left( \min \limits_{\pi \in \Pi} -H(\pi) + \mathbb{E}_{\pi}[\lambda d(s,a) -r(s,a)] \right) \\
& -\mathbb{E}_{\pi_{E}}[\lambda d(s,a) -r(s,a)]+\psi(r) \\
\end{aligned}
\end{equation}

Given $(\tilde{r},\tilde{\lambda}) \in IRL_\psi(\pi_E,d)$, our objective is to learn a policy defined by $RL(\tilde{r},\tilde{\lambda})$. To characterize $RL(\tilde{r},\tilde{\lambda})$, it is commonly beneficial to convert optimization problems involving policies into convex problems. We use occupancy measure $\rho_\pi$ to accomplish this. After which we express the expected value of the reward and the expected value of the constraint as: $\mathbb{E}_{\pi}[r(s,a)]=\sum_{s,a} \rho_\pi(s,a)r(s,a)$ and $\mathbb{E}_{\pi}[d(s, a)]=\sum_{s, a} \rho_\pi(s, a)d(s, a)$ as described in \citet{altman1999constrained}. IRL can be reformulated as:
\begin{equation}
\label{IRL regu 1}
\begin{aligned}
& IRL_\psi(\pi_E,d)= \mathop{\arg\max} \limits_{r \in \mathbb{R}^{\mathcal{S}\times \mathcal{A}} \atop \lambda \geq 0} \min \limits_{\pi \in \Pi}  -H(\pi) +\psi(r) + \\
& \sum_{s,a}(\rho_\pi(s,a)-\rho_{\pi_E}(s,a)) [\lambda d(s,a) - r(s,a)] \\
\end{aligned}
\end{equation}

We then characterize $RL(\tilde{r},\tilde{\lambda})$, the policy learned by RL on the reward recovered by IRL as the optimal solution of Eq.\eqref{CMDP gail obj}.

\begin{proposition}
\label{proposition 1}
(Theorem 2 of \citet{syed2008apprenticeship}) If $\rho \in \mathcal{D}$, then $\rho$ is the occupancy measure for $\pi_\rho(a|s) \triangleq \rho(s,a)/\sum_a' \rho(s,a')$, and $\pi_\rho$ is the only policy whose occupancy measure is $\rho$.
\end{proposition}

\begin{proposition}
\label{proposition 2}
(Lemma 3.1 of \citet{ho2016generative}) Let $\bar{H}(\rho) = -\sum_{s,a}\rho(s,a) \log(\rho(s,a)/\sum_{a'} \rho(s,a'))$. Then, $\bar{H}$ is strictly concave, and for all $\pi \in \Pi$ and $\rho \in \mathcal{D}$, we have $H(\pi)=\bar{H}(\rho_\pi)$ and $\bar{H}(\rho)=H(\pi_\rho)$.
\end{proposition}

\begin{proposition}
\label{proposition 3}
Let $(\tilde{r},\tilde{\lambda})  \in IRL_\psi(\pi_E,d)$, $\tilde{\pi} \in RL(\tilde{r},\tilde{\lambda})$, and 
\begin{equation}
\begin{aligned}
&\pi_A \in \mathop{\arg\min} \limits_{ \pi}  -H(\pi) + \psi^*(\rho_\pi-\rho_{\pi_E})\\
&= \mathop{\arg\min} \limits_{ \pi} \max \limits_{r,\lambda} -H(\pi) +\psi(r) + \\
& \sum_{s,a}(\rho_\pi(s,a) - \rho_{\pi_E}(s,a))[\lambda d(s,a) - r(s,a)] \\
\end{aligned}
\end{equation}
Then $\pi_A = \tilde{\pi}$.
\end{proposition}

\textit{Proof.} Let $\rho_A$ be the occupancy measure of $\pi_A$ and $\tilde{\rho}$ be the occupancy measure of $\tilde{\pi}$. By using Proposition \ref{proposition 1}, we define $\bar{L}:\mathcal{D}\times \mathbb{R}^{\mathcal{S}\times \mathcal{A}} \times \mathbb{R} \to \mathbb{R}$ by
\begin{equation}
\label{L}
\begin{aligned}
& \bar{L}(\rho,(r,\lambda)) = -\bar{H}(\rho) + \psi(r) + \\
& \sum_{s,a}(\rho_\pi(s,a) - \rho_{\pi_E}(s,a))[\lambda d(s,a) - r(s,a)] 
\end{aligned}
\end{equation}

The following relationship then holds:
\begin{equation}
\label{P2}
\rho_A \in \mathop{\arg\min} \limits_{\rho \in \mathcal{D}} \max \limits_{r,\lambda} \bar{L}(\rho,(r,\lambda))
\end{equation}

\begin{equation}
\label{P3}
(\tilde{r},\tilde{\lambda}) \in \mathop{\arg\max} \limits_{r,\lambda} \min \limits_{\rho \in \mathcal{D}} \bar{L}(\rho,(r,\lambda))
\end{equation}

\begin{equation}
\label{P4}
\tilde{\rho} \in \mathop{\arg\min} \limits_{\rho \in \mathcal{D}} \bar{L}(\rho,(\tilde{r},\tilde{\lambda}))
\end{equation}

$\mathcal{D}$ is compact and convex, $\mathbb{R}^{\mathcal{S}\times \mathcal{A}}$ is convex. Due to convexity of $-\bar{H}$,it follows that $\bar{L}(\rho,\cdot)$ is convex for all $\rho$. $\bar{L}(\cdot,(r,\lambda))$ is concave for all $(r,\lambda)$ (see proof in \ref{Other 1}).

Therefore, we can use minimax duality \cite{millar1983minimax}:
\begin{equation}
\label{P5}
\min \limits_{\rho \in \mathcal{D}} \max \limits_{r \in \mathcal{R} \atop \lambda} \bar{L}(\rho,(c,\lambda)) = \max \limits_{r \in \mathcal{R} \atop \lambda} \min \limits_{\rho \in \mathcal{D}} \bar{L}(\rho,(c,\lambda))
\end{equation}

Hence, from Eqs.(\ref{P2}) and (\ref{P3}), $(\rho_A,(\tilde{r},\tilde{\lambda}))$ is a saddle point of $\bar{L}$, which implies that:
\begin{equation}
\label{P6}
\rho_A \in \mathop{\arg\min} \limits_{\rho \in \mathcal{D}} \bar{L}(\rho,(\tilde{r},\tilde{\lambda}))
\end{equation}
Because $\tilde{L}(\cdot,(r,\lambda))$ is strictly concave for all $(r,\lambda)$, Eqs.(\ref{P4}) and (\ref{P6}) imply $\rho_A = \tilde{\rho} $. Since policies whose corresponding occupancy measure are unique (Proposition \ref{proposition 2}), finally we get $\pi_A = \tilde{\pi}$.

Proposition \ref{proposition 3} illustrates the process of IRL in finding the optimal reward function and Lagrangian multiplier, represented by ($r^*,\lambda^*$). By utilizing the output of IRL, reinforcement learning can be executed to obtain the optimal policy, represented by $\pi^*$. And we prove that $\pi^*$ is the same as by directly solving the $\psi$-regularized imitation learning problem $\tilde{L}$. Furthermore, $\psi$-regularized imitation learning aims to identify a policy whose occupancy measure is similar to that of an expert, as measured by the convex function $\psi^*$. Subsequently, we deduce the form of $\psi^*$.

\begin{algorithm}[!tb]
\caption{Meta-Gradients for Lagrangian Multipliers }
\label{Algorithm MAML}
\textbf{Input}: initial parameters of policy $\theta$, reward value network $\phi_r$, cost value network $\phi_d$, discriminator network $\omega$, batch size $K$, a set of expert trajectories $\Phi_E = \{\tau_E \sim \pi_E\}$, initial Lagrangian multipliers $\lambda$, entropy parameter $\beta$, learning rates $\alpha_r,\alpha_d,\alpha_\omega,\alpha_\lambda$ \\
\textbf{Output}: Optimal policy $\pi_\theta$ 
\begin{algorithmic}[1] 
{\fontsize{9.5pt}{10.5pt} \selectfont
\STATE Compute the average cost of expert trajectories: $J_\mathbb{E}= \frac{1}{|\Phi_k|}\sum_{\tau \in \Phi_E} \sum_{t=1}^T d_t$
\FOR {$k = 1,2,...$}
\STATE Collect set of learner's trajectories $\Phi_k = \{\tau_i \}$ by running policy $\pi_{\theta_k}$ for $K$ time steps.
\STATE Collect the reward $r_t$ of $K$ time steps by using the discriminator output:$r_t = -\log (D_\omega(s_t,a_t))$ 
\STATE Compute $V^r_{\phi_r}(s_t)$ and $V^d_{\phi_d}(s_t)$ of $K$ time steps.
\STATE Compute the reward and cost advantage $A^r(s_t,a_t)$ and $A^d(s_t,a_t)$, reward to go $\hat{R}^r_t$ and cost to go $\hat{R}^d_t$ of $K$ time steps by using GAE.
\STATE Compute the average episode cost of learner's trajectories: $J_k= \frac{1}{|\Phi_k|} \sum_{\tau \in \Phi_k} \sum_{t=1}^T d_t$
\STATE Split the data of $K$ time steps into training and validation sets $K_{tr},K_{va}$
\STATE \textbf{Inner loss:}
\STATE Update policy by using TRPO rule: \\
\resizebox{0.9\columnwidth}{!}{$ \theta' = \arg \max \limits_{\theta} \sum_{t=1}^{K_{tr}} \frac{\pi_{\theta}(a_t|s_t)}{\pi_{\theta_k}(a_t|s_t)}(A^r(s_t,a_t)-\lambda A^d(s_t,a_t)) + \beta H(\pi_{\theta_k})$}
\STATE Update reward value network:\\
       $\phi_r' \gets \phi_r -  \frac{1}{K_{tr}}  \sum_{t=1}^{K_{tr}} \alpha_r \bigtriangledown_{\phi_r} (V^r_{\phi_r}(s_t)-\hat{R}^r_t)^2 $
\STATE Update cost value network:\\
       $\phi_d' \gets \phi_d -  \frac{1}{K_{tr}}  \sum_{t=1}^{K_{tr}} \alpha_d\alpha_r  \bigtriangledown_{\phi_d} (V^d_{\phi_d}(s_t)-\hat{R}^d_t)^2 $
\STATE Update discriminator network: \\
       $\omega' \gets \omega +  \frac{1}{K} \sum_{t=1}^K \alpha_\omega (\bigtriangledown_{\omega} [\log(D_{\omega}(s_t,a_t))] + $\\
       $\bigtriangledown_{\omega} [\log(1- D_{\omega}(s_t,a_t)] )  $
\STATE Update Lagrangian multipliers: \\
       $\lambda' \gets \lambda + \alpha_\lambda (J_k-J_\mathbb{E})$

\STATE \textbf{Outer loss:}
\STATE Meta-parameter update: \\
       $\lambda'' \gets \lambda' - \frac{1}{K_{va}}  \sum_{t=1}^{K_{va}} \bigtriangledown_{\lambda'} (A^r(s_t,a_t)- \lambda' d_t)^2 $
       
\STATE $ \theta \gets \theta'$, $\phi_r \gets \phi_r'$, $\phi_d\gets \phi_d'$, $\omega \gets \omega'$, $ \lambda \gets \lambda''$. 

\ENDFOR 
}
\end{algorithmic}
\end{algorithm}

\begin{algorithm}[htb]
\caption{Cost Violation based Alternating Gradient}
\label{Algorithm BVF}
\textbf{Input}: initial parameters of policy $\theta$, reward value network $\phi_r$, cost value network $\phi_d$, discriminator network $\omega$, batch size $K$, a set of expert trajectories $\Phi_E = \{\tau_E \sim \pi_E\}$, entropy parameter $\beta$, learning rates $\alpha_r,\alpha_d,\alpha_\omega$.\\
\textbf{Output}: Optimal policy $\pi_\theta$ 
\begin{algorithmic}[1] 
{\fontsize{9.5pt}{10.5pt} \selectfont
\STATE Compute the average cost of expert trajectories: $J_\mathbb{E}= \frac{1}{|\Phi_k|}\sum_{\tau \in \Phi_E} \sum_{t=1}^T d_t$
\FOR {$k = 1,2,...$}
\STATE Collect set of learner's trajectories $\Phi_k = \{\tau_i \}$ by running policy $\pi_{\theta_k}$ for $K$ time steps.
\STATE Collect the reward $r_t$ of $K$ time steps by using the discriminator output:$r_t = -\log (D_\omega(s_t,a_t))$ 
\STATE Compute $V^r_{\phi_r}(s_t)$ and $V^d_{\phi_d}(s_t)$ of $K$ time steps.
\STATE Compute the reward and cost advantage $A^r(s_t,a_t)$ and $A^d(s_t,a_t)$, reward to go $\hat{R}^r_t$ and cost to go $\hat{R}^d_t$ of $K$ time steps by using GAE.
\STATE Compute the average episode cost of learner's trajectories: $J_k= \frac{1}{|\Phi_k|} \sum_{\tau \in \Phi_k} \sum_{t=1}^T d_t$
\IF{$J_k \leq J_\mathbb{E}$}
\STATE Update policy towards maximizing the return: \\
\resizebox{0.8\columnwidth}{!}{$\theta' = \arg \max \limits_{\theta} \sum_{t=1}^K \frac{\pi_{\theta}(a_t|s_t)}{\pi_{\theta_k}(a_t|s_t)}A^r(s_t,a_t) + \beta H(\pi_{\theta_k})$}
\ELSE
\STATE Update policy towards minimizing the cost: \\
\resizebox{0.8\columnwidth}{!} {$\theta' = \arg \min \limits_{\theta} \sum_{t=1}^K \frac{\pi_{\theta}(a_t|s_t)}{\pi_{\theta_k}(a_t|s_t)}A^d(s_t,a_t) -  \beta H(\pi_{\theta_k})$}
\ENDIF
\STATE Update reward value network:\\
       $\phi_r' \gets \phi_r -  \frac{1}{K}  \sum_{t=1}^K \alpha_r \bigtriangledown_{\phi_r} (V^r_{\phi_r}(s_t)-\hat{R}^r_t)^2 $
\STATE Update cost value network:\\
       $\phi_d' \gets \phi_d -  \frac{1}{K}  \sum_{t=1}^K \alpha_d \bigtriangledown_{\phi_d} (V^d_{\phi_d}(s_t)-\hat{R}^d_t)^2 $
\STATE Update discriminator network: \\
       $\omega' \gets \omega +  \frac{1}{K} \sum_{t=1}^K \alpha_\omega (\bigtriangledown_{\omega} [\log(D_{\omega}(s_t,a_t))] + $\\
       $\bigtriangledown_{\omega} [\log(1- D_{\omega}(s_t,a_t)] )  $
\STATE $ \theta \gets \theta'$, $\phi_r \gets \phi_r'$, $\phi_d\gets \phi_d'$, $\omega \gets \omega'$.
\ENDFOR 
}
\end{algorithmic}
\end{algorithm}

\subsection{Step 2}

In their GAIL paper, \citet{ho2016generative} present a cost regularizer, $\psi_{GA}$, that leads to an imitation learning algorithm, as outlined in Eq.(\ref{ILobj2}), which aims to minimize the Jensen-Shannon divergence between the occupancy measures. Specifically, they convert a surrogate loss function, $\phi$, which is used for binary classification of state-action pairs drawn from the occupancy measures $\rho_\pi$ and $\rho_{\pi_E}$, into cost function regularizers $\phi$, such that $\phi^*(\rho_\pi-\rho_{\pi_E})$ represents the minimum expected risk, $R_\phi(\rho_\pi,\rho_{\pi_E})$, for the function $\phi$ \cite{ho2016generative}.

\begin{equation}
 \label{expected risk}
  R_\phi(\rho_\pi,\rho_{\pi_E}) = \sum_{s,a} \max \limits_{\gamma \in \mathbb{R}}  \rho_\pi(s,a)\phi(\gamma) + \rho_{\pi_E}(s,a) \phi(-\gamma)    
\end{equation}

Here we use the same formula of surrogate loss function $\phi$ as in GAIL paper: $\psi_\phi(c) = \sum_{\rho_{\pi_E}}g_\phi(c(s,a))$, where $g_\phi(x) = -x + \phi(-\phi^{-1}(-x))$, $\phi$ is a strictly decreasing convex function (Proposition \ref{proposition 1} from \citet{ho2016generative}). Noted that in GAIL paper they adopt cost function $c(s,a)$ not reward function $r(s,a)$, then we write in this form: $\psi_\phi(-r) = \sum_{\rho_{\pi_E}}g_\phi(-r(s,a))$.

Then formulation of $\psi_\phi^*(\rho_\pi - \rho_{\pi_E})$ is represented as follows(see proof in \ref{Other 2}):

\begin{equation}
\label{regu 1}
\begin{aligned}
&\psi_\phi^*(\rho_\pi-\rho_{\pi_E})\\
& = -R_\phi(\rho_\pi,\rho_{\pi_E}) + \max \limits_{\lambda} \sum_{s,a} \lambda(\rho_\pi(s,a)-\rho_{\pi_E}(s,a)) d(s,a) \nonumber \\
\end{aligned}
\end{equation}

Using the logistic loss $\phi(\gamma) = \log (1 + e^{-\gamma})$, the same form in GAIL paper, then $-R_\phi(\rho_\pi,\rho_{\pi_E}) = \max \limits_{ D  \in (0,1)^{\mathcal{S}\times \mathcal{A}}}  \sum_{s,a} \rho_\pi(s,a) \log D(s,a)+ \rho_{\pi_E}(s,a) \log(1- D(s,a)) $. Therefore, we obtain the final form of $\psi^*(\rho_\pi - \rho_{\pi_E})$ as follows: 
\begin{equation}
\label{regu 2}
\begin{aligned}
& \psi^*(\rho_\pi-\rho_{\pi_E}) = \max \limits_{ D  \in (0,1)^{\mathcal{S}\times \mathcal{A}} \atop \lambda}  \sum_{s,a}  \rho_\pi(s,a) \log D(s,a) +\\
& \rho_{\pi_E}(s,a) \log(1- D(s,a)) + \lambda (\rho_\pi(s,a)-\rho_{\pi_E}(s,a)) d(s,a) \\
& = \max \limits_{ D \in (0,1)^{\mathcal{S}\times \mathcal{A}} \atop \lambda}  \mathbb{E}_\pi[\log D(s,a)]+ \mathbb{E}_{\pi_E}[\log(1- D(s,a))] \\
& + \lambda (\mathbb{E}_\pi[d(s,a)]-\mathbb{E}_{\pi_E}[d(s,a)]) \nonumber \\
\end{aligned}
\end{equation}

\subsection{Other Proofs}
\subsubsection{Prove concavity of $\bar{L}$}
\label{Other 1}
$\bar{L}(\cdot,(r,\lambda))$ is concave for all $(r,\lambda)$.
\textit{Proof} We known that $\psi(r)$ is concave, suppose $\alpha\in[0,1]$.
\begin{equation}
\label{L concave}
\begin{aligned}
&\bar{L}(\cdot,(\alpha r_1+(1-\alpha) r_2,\alpha \lambda_1 + (1-\alpha) \lambda_2)) = -\bar{H}(\rho) + \\
& \psi(\alpha r_1 +(1-\alpha)r_2) +\\
&\sum_{s,a}(\rho_\pi - \rho_{\pi_E}) [d(\alpha \lambda_1+(1-\alpha)\lambda_2) - (\alpha r_1+(1-\alpha) r_2)]\\
& \geq \alpha \psi(r_1) + (1-\alpha) \psi(r_2) + \alpha \sum_{s,a}(\rho_\pi - \rho_{\pi_E}) (\lambda_1 d - r_1)\\
& + (1-\alpha) \sum_{s,a}(\rho_\pi - \rho_{\pi_E}) (\lambda_2 d - r_2) \nonumber \\
\end{aligned}
\end{equation}
Therefore, $\bar{L}(\cdot,(\alpha r_1+(1-\alpha) r_2,\alpha \lambda_1 + (1-\alpha) \lambda_2)) \geq \bar{L}(\cdot,(\alpha r_1,\alpha \lambda_1 ) + \bar{L}(\cdot,((1-\alpha) r_2,(1-\alpha) \lambda_2)) $, $\bar{L}(\cdot,(r,\lambda))$ is concave for all $(r,\lambda)$.

\subsubsection{Proof of $\psi_\phi^*(\rho_\pi - \rho_{\pi_E})$}
\label{Other 2}
We deduce the form of $\psi_\phi^*(\rho_\pi - \rho_{\pi_E})$ as:
\begin{equation}
\begin{aligned}
& \psi_\phi^*(\rho_\pi-\rho_{\pi_E}) = \\
& -R_\phi(\rho_\pi,\rho_{\pi_E}) + \max \limits_{\lambda} \lambda \sum_{s,a}(\rho_\pi(s,a)-\rho_{\pi_E}(s,a)) d(s,a) \nonumber \\
\end{aligned} 
\end{equation}
 We will simplify notation by using the symbols $\rho_\pi$, $\rho_{\pi_E}$, $r$, and $d$ to represent $\rho_\pi(s,a)$,$\rho_{\pi_E}(s,a)$,$r(s,a)$ and $d(s,a)$, respectively.

\begin{equation}
\label{regu}
\begin{aligned}
&\psi_\phi^*(\rho_\pi-\rho_{\pi_E}) = \max \limits_{r \in \mathcal{R} \atop \lambda} \sum_{s,a}(\rho_\pi-\rho_{\pi_E})(\lambda d - r) -\psi_\phi(-r) \\  
& = \max \limits_{r \in \mathcal{R} \atop \lambda} \sum_{s,a}(\rho_\pi-\rho_{\pi_E})(\lambda d-r) - \sum_{s,a}\rho_{\pi_E}  g_\phi(-r) \\
& = \max \limits_{r \in \mathcal{R}}  \sum_{s,a}(\rho_\pi-\rho_{\pi_E})(-r) - \sum_{s,a}\rho_{\pi_E} (r +\phi(-\phi^{-1}(r)) ) \\
& + \max \limits_{\lambda} \sum_{s,a} \lambda(\rho_\pi -\rho_{\pi_E} )d  \\
& =\max \limits_{r \in \mathcal{R}} \sum_{s,a}\rho_\pi(-r) - \sum_{s,a}\rho_{\pi_E} \phi(-\phi^{-1}(r)) \\
& + \max \limits_{\lambda} \sum_{s,a} \lambda(\rho_\pi -\rho_{\pi_E} )d \nonumber \\
\end{aligned}
\end{equation}

Then we make the change of variables $r \rightarrow \phi(\gamma)$, the above equation becomes:
\begin{equation}
\label{regu 1 prove}
\begin{aligned}
&\psi_\phi^*(\rho_\pi-\rho_{\pi_E}) = \\
& \sum_{s,a} \max \limits_{\gamma \in \mathbb{R}}  \rho_\pi(-\phi(\gamma)) - \rho_{\pi_E}\phi(-\gamma) + \max \limits_{\lambda} \lambda \sum_{s,a}(\rho_\pi-\rho_{\pi_E}) d \\
& = -R_\phi(\rho_\pi,\rho_{\pi_E}) + \max \limits_{\lambda} \lambda \sum_{s,a}(\rho_\pi-\rho_{\pi_E}) d \nonumber \\
\end{aligned}
\end{equation}
Therefore, we prove Theorem \ref{Theo 1} and the objective function of cost-constrained imitation learning is Eq.(\ref{obj2}).

\section{B. Algorithms for MALM and CVAG}
Algorithms \ref{Algorithm MAML} and \ref{Algorithm BVF} are pseudocode for Meta-Gradients for Lagrangian multipliers (MALM) and Cost Violation based Alternating Gradient (CVAG) methods.

\section{C. Experiment Figures}
\subsection{C.1 Experiments results }
\begin{figure*}[!htb]
  \centering
  \includegraphics[width=7in]{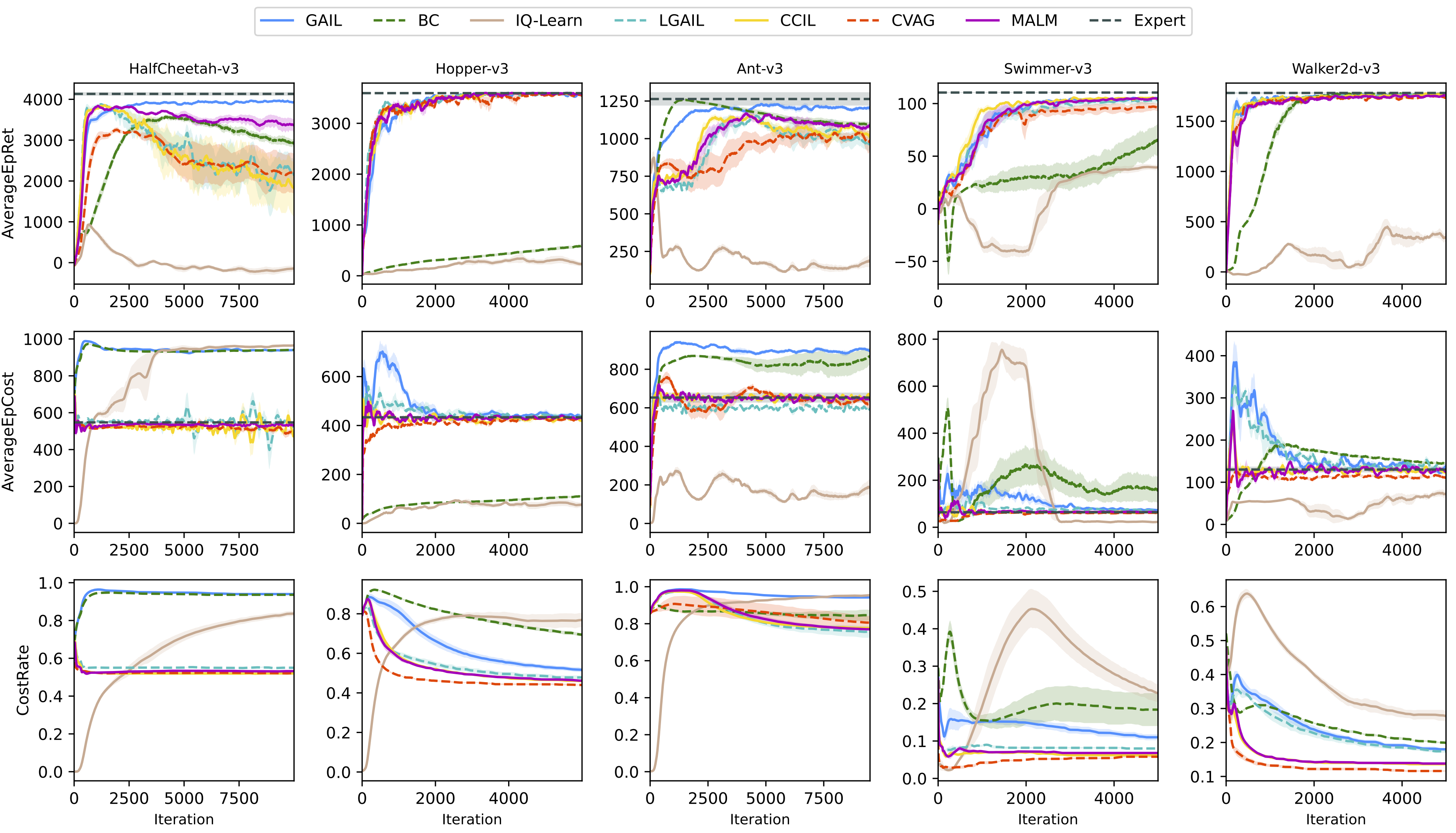}
  \caption{Performance of MuJoCo environments. The x-axes indicate the number of iterations, and the y-axes indicate the performance of the agent, including average rewards/costs/cost rates with standard deviations.}
  \label{Result Mujoco}
\end{figure*}

\begin{figure*}[!htb]
  \centering
  \includegraphics[width=7in]{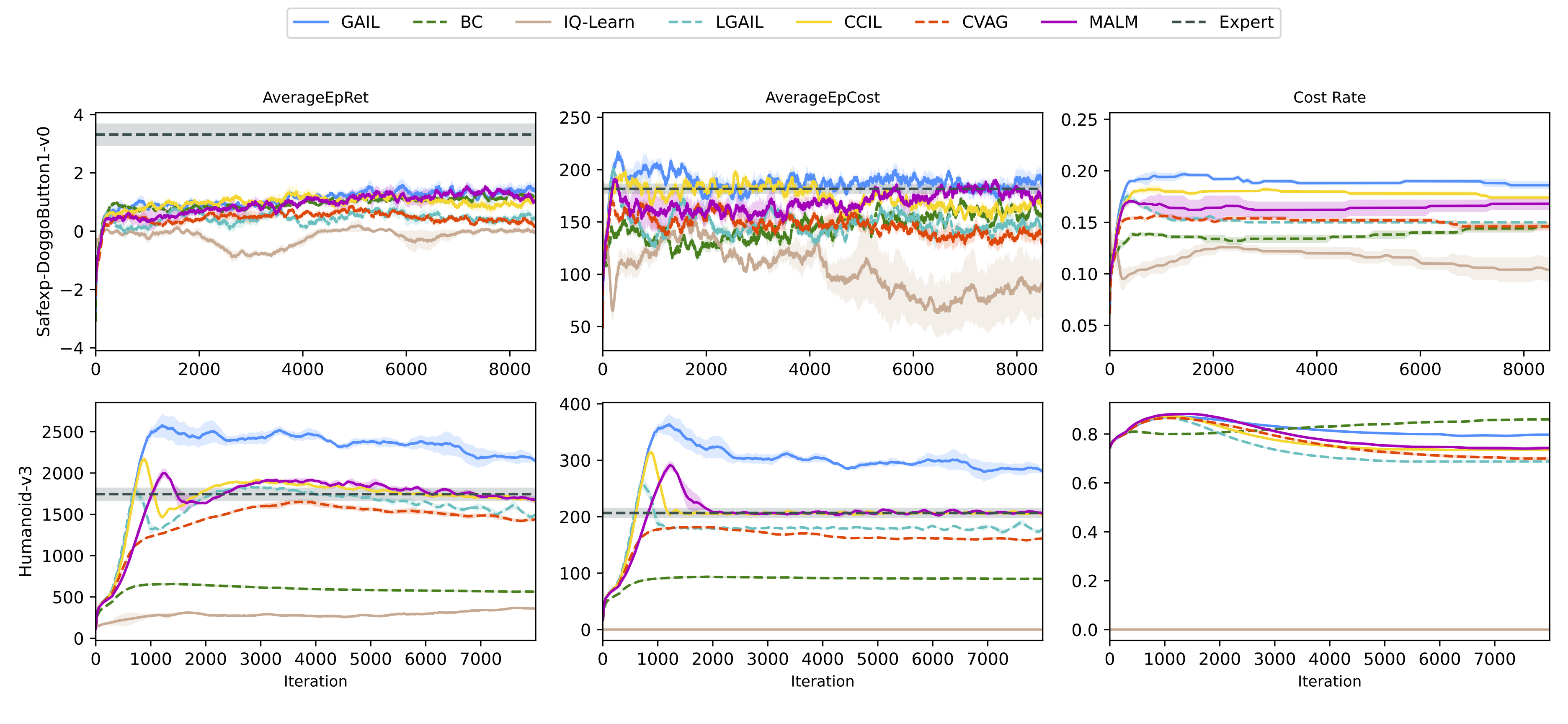}
  \caption{Performance of Humanoid and DoggoButton tasks The x-axes indicate the number of iterations, and the y-axes indicate the performance of the agent, including average rewards/costs/cost rates with standard deviations.}
  \label{Result Humanoid}
\end{figure*}

Figures \ref{Result Mujoco} and \ref{Result Humanoid} are experiment results from MuJoCo, Humanoid, and DoggoButton tasks. 

\subsection{C.2 Experiment Hyperparameters}
Table \ref{experiment hyperparameters} lists hyperparameters we used in our experiments.
\begin{table}[!htb]
    \centering
{\fontsize{9.5pt}{10.5pt} \selectfont
    \begin{tabular}{cc}
        \toprule
        \small Hyperparameter & Value \\
        \midrule
       Policy and Value network size & (100,100) \\
       Actor and Critic network size (for IQ-Learn) & (256,256) \\
       Activation & $\tanh$ \\
       Batch Size  & 2000 \\
       Generator network update times & 3 \\
       Discriminator network update times & 1\\
       Generalized Advantage Estimation $\gamma$ & 0.995\\
       Generalized Advantage Estimation $\lambda$ & 0.97\\
       Maximum KL & 0.01\\
       Learning rate (Value network) & 1 $\times$  $10 ^ {-3}$ \\
       Learning rate (Discriminator network) & 3 $\times$  $10 ^ {-4}$ \\
       Policy entropy & 0.0 \\
       Discriminator entropy &  1 $\times$  $10 ^ {-3}$  \\
       Initial Lagrangian penalty & 0.01 \\
       Lagrangian penalty learning rate & 0.05 \\
       Meta-learning rate & 0.05 \\
       \bottomrule
    \end{tabular}
}
    \caption{Hyperparameters used in experiments.}
    \label{experiment hyperparameters}
\end{table}

\end{appendices}

\end{document}